\documentstyle[aclap,lingmacros,times]{article}

\title{PARADISE: A Framework for Evaluating Spoken Dialogue Agents}

\author{Marilyn A. Walker, Diane J. Litman, Candace A. Kamm and Alicia Abella\\
AT\&T Labs---Research \\
180 Park Avenue \\
Florham Park, NJ 07932-0971 USA \\
walker,diane,cak,abella@research.att.com }

\begin{document}
\input{psfig}
\maketitle
\bibliographystyle{fullname}

\begin{abstract}

This paper presents PARADISE (PARAdigm for DIalogue System
Evaluation), a general framework for evaluating spoken dialogue
agents.  The framework decouples task requirements from an agent's
dialogue behaviors, supports comparisons among dialogue strategies,
enables the calculation of performance over subdialogues and whole
dialogues, specifies the relative contribution of various factors to
performance, and makes it possible to compare agents performing
different tasks by normalizing for task complexity.
\end{abstract}

\section{Introduction}
\label{intro-sec}

Recent advances in dialogue modeling, speech recognition, and natural
language processing have made it possible to build spoken dialogue
agents for a wide variety of applications.\footnote{We use the term
agent to emphasize the fact that we are evaluating a speaking entity
that may have a personality. Readers who wish to may substitute the
word ``system'' wherever ``agent'' is used.} Potential benefits of such
agents include remote or hands-free access, ease of use, naturalness,
and greater efficiency of interaction.  However, a critical obstacle
to progress in this area is the lack of a general framework for
evaluating and comparing the performance of different dialogue agents.

One widely used approach to evaluation is based on the notion of a
reference answer~\cite{HirschmanEtAl90}. An agent's responses to a
query are compared with a predefined key of minimum and maximum
reference answers; performance is the proportion of responses that
match the key.  This approach has many widely acknowledged limitations~\cite{HirschmanPao93,DanieliEtAl92,BatesAyuso}, e.g.,
although there may be many potential dialogue strategies for carrying
out a task, the key is tied to one particular dialogue strategy.

In contrast, agents using different dialogue strategies can be
compared with measures such as inappropriate utterance ratio, turn
correction ratio, concept accuracy, implicit recovery and transaction
success~\cite{DanieliGerbino95,HirschmanPao93,PHSZ92,SimpsonFraser93,SWP92}.
Consider a comparison of two train timetable information
agents~\cite{DanieliGerbino95}, where Agent A in Dialogue 1 uses an
explicit confirmation strategy, while Agent B in Dialogue 2 uses an
implicit confirmation strategy:

\enumsentence{ User: I want to go from Torino
to Milano.  \\ Agent A: Do you want to go from Trento to Milano?
Yes or No?  \\ User: No.  }

\enumsentence{ User: I want to travel from
Torino  to Milano. \\ Agent B: At which time do you want to leave from
Merano to Milano? \\ User: No, I want to leave from Torino in the
evening. }

\noindent
Danieli and Gerbino found that Agent A had a higher transaction
success rate and produced less inappropriate and repair utterances
than Agent B, and thus concluded that Agent A was more robust than
Agent B.

However, one limitation of both this approach and the reference answer
approach is the inability to generalize results to other tasks and
environments~\cite{Fraser95}.  Such generalization requires the
identification of factors that affect
performance~\cite{Cohen95,sjg96}.  For example, while Danieli and
Gerbino found that Agent A's dialogue strategy produced dialogues that
were approximately twice as long as Agent B's, they had no way of
determining whether Agent A's higher transaction success or Agent B's
efficiency was more critical to performance.  In addition to agent
factors such as dialogue strategy, task factors such as database size
and environmental factors such as background noise may also be
relevant predictors of performance.

These approaches are also limited in that they currently do not
calculate performance over subdialogues as well as whole dialogues,
correlate performance with an external validation criterion, or
normalize performance for task complexity.

\begin{figure}[htb]
\centerline{\psfig{figure=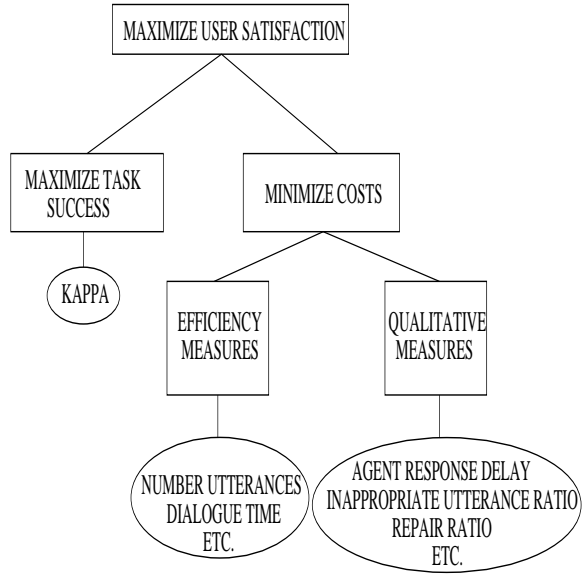,height=3.0in,width=3.0in}}
\caption{PARADISE's structure of objectives for spoken dialogue performance}
\label{objectives-fig}
\end{figure}

This paper describes PARADISE, a general framework for evaluating
spoken dialogue agents that addresses these limitations.  PARADISE
supports comparisons among dialogue strategies by providing a task
representation that decouples {\it what} an agent needs to achieve in
terms of the task requirements from {\it how} the agent carries out
the task via dialogue.  PARADISE uses a decision-theoretic framework
to specify the relative contribution of various factors to an agent's
overall {\it performance}.  Performance is modeled as a weighted
function of a task-based success measure and dialogue-based cost
measures, where weights are computed by correlating user satisfaction
with performance.  Also, performance can be calculated for
subdialogues as well as whole dialogues. Since the goal of this paper
is to explain and illustrate the application of the PARADISE
framework, for expository purposes, the paper uses simplified domains
with hypothetical data throughout.  Section 2 describes PARADISE's
performance model, and Section 3 discusses its generality, before
concluding in Section 4.

\section{A Performance Model for Dialogue}
\label{perf-sec}

PARADISE uses methods from decision
theory~\cite{KeeneyRaiffa76,Doyle92} to combine a disparate set of
performance measures (i.e., user satisfaction, task success, and
dialogue cost, all of which have been previously noted in the
literature) into a single performance evaluation function.  The use of
decision theory requires a specification of both the objectives of the
decision problem and a set of measures (known as attributes in
decision theory) for operationalizing the objectives. The PARADISE
model is based on the structure of objectives (rectangles) shown in
Figure~\ref{objectives-fig}.  The PARADISE model posits that
performance can be correlated with a meaningful external criterion such as
usability, and
thus that the overall goal of a spoken dialogue agent is to maximize
an objective related to usability. User
satisfaction ratings~\cite{Kamm95,SWP92,PHSZ92} have been
frequently used in the literature as an external indicator of the
usability of a dialogue agent.  The model further posits that two
types of factors are potential relevant contributors to user
satisfaction (namely task success and dialogue costs), and that two types of factors are
potential relevant contributors to costs~\cite{Walker96a}.

In addition to the use of decision theory to create this objective structure,
other novel aspects of PARADISE include the use of
the Kappa coefficient~\cite{Carletta96,Siegel88} to operationalize
task success, and the use of linear regression to
quantify the relative contribution of the success and cost factors to user satisfaction.

The remainder of this section explains the measures (ovals in
Figure 1) used to operationalize the set of objectives, and the
methodology for estimating a quantitative performance function that
reflects the objective structure.  Section~\ref{avm-sec} describes
PARADISE's task representation, which is needed to calculate the
task-based success measure described in Section~\ref{succ-sec}.
Section~\ref{cost-sec} describes the cost measures considered in
PARADISE, which reflect both the efficiency and the naturalness of an
agent's dialogue behaviors. Section~\ref{perf-func-sec} describes the
use of linear regression and user satisfaction to estimate the
relative contribution of the success and cost measures in a single performance
function.
Finally, Section~\ref{subdialogues-sec} explains how performance
can be calculated for subdialogues as well as whole
dialogues, while Section~\ref{summary-sec}  summarizes the method.

\subsection{Tasks as Attribute Value Matrices}
\label{avm-sec}

A general evaluation framework requires a task representation that
decouples {\it what} an agent and user accomplish from {\it how} the
task is accomplished using dialogue strategies.  We propose that an
{\it attribute value matrix (AVM)} can represent many dialogue
tasks. This consists of the information that must be exchanged between
the agent and the user during the dialogue, represented as a set of
ordered pairs of attributes and their possible values.\footnote{For
infinite sets of values, actual values found in the experimental data constitute
the required finite set.}

As a first illustrative example, consider a simplification of the
train timetable domain of Dialogues 1 and 2, where the timetable only
contains information about rush-hour trains between four cities, as
shown in Table 1.  This AVM consists of four attributes (abbreviations
for each attribute name are also shown).\footnote{The AVM serves as an
evaluation mechanism only.  We are not claiming that AVMs determine an
agent's behavior or serve as an utterance's semantic representation.}
In Table 1, these attribute-value pairs are annotated with the
direction of information flow to represent who acquires the
information, although this information is not used for evaluation.  During the dialogue the agent must acquire from the user
the values of DC, AC, and DR, while the user must acquire DT.

{\scriptsize
\begin{table}[htb]
\begin{center}
\begin{tabular}{|l|l|c|} \hline
attribute  &possible values & information flow\\ \hline
depart-city (DC) &Milano, Roma, Torino, Trento & to agent \\
arrival-city (AC) &Milano, Roma, Torino, Trento & to agent \\
depart-range (DR) &morning,evening & to agent\\
depart-time (DT) &6am,8am,6pm,8pm & to user\\ \hline
\end{tabular}
\label{aclAVM}
\caption{Attribute value matrix, simplified train timetable domain}
\end{center}
\end{table} }

{\scriptsize
\begin{figure}[htb]
\begin{center}
\begin{tabular}{rll}
A1:	&Hello, This is Train Enquiry Service. &  DC,AC,DR,DT \\
&	Please speak after the tone.  & DC,AC,DR,DT \\
&	Which information do you need?  & DC,AC,DR,DT \\
U1:&	I want to go from Torino to Milano.  & DC,AC\\
A2:&	Do you want to go from Trento to Milano?  & DC,AC \\
	&Yes or No?  & DC,AC \\
U2:	&No.  & DC,AC \\
A3:&	Do you want to leave from Trento? & DC\\
&	Yes or No?  & DC \\
U3:&	No.  & DC \\
A4:&	Where do you want to leave from? & DC \\
U4:&	I want to leave from Torino. & DC \\
A5:&	Do you want to leave from Torino? & DC \\
&	Yes or No?  & DC \\
U5:&	Yes.  & DC \\
A6:&	Do you want to go to Milano?& AC \\
U6:&	Yes.& AC \\
A7:&	At which time do you want to leave?& DR\\
U7:&	I want to travel in the evening.& DR \\
A8:&	Do you want to leave between 6 and 9 p.m.?& DR \\
&	Yes or No?& DR \\
U8:&	Yes.& DR \\
A9:&	There is a train leaving at 8:00  p.m. &DT\\
&	\ldots
\end{tabular}
\label{d1}
\caption{Agent A dialogue interaction (Danieli and Gerbino, 1995)}
\end{center}
\end{figure}
}
{\scriptsize
\begin{figure}[ht]
\begin{center}
\begin{tabular}{rll}
B1:	&Hello, This is Train Enquiry Service.  & DC,AC,DR,DT \\
	&Please speak after the tone.& DC,AC,DR,DT \\
	&Which information do you need?& DC,AC,DR,DT \\
U1:	&I want to travel from Torino to Milano.& DC,AC\\
B2:	&At which time do you want to leave & \\
	& from Merano to Milano?& DC,AC,DR \\
U2:	&No, I want to leave from Torino \\
	&in the evening.& DC, DR \\
B3:	&Do you want to leave from Torino  & \\
	& between 6 and 11 p.m?  & DC, DR \\
	&Please answer Yes or No. & DC, DR \\
U3:	&Yes.& DC, DR \\
B4:	&A train leaves at 8 p.m.& DT\\
	&	\ldots
\end{tabular}
\label{d2}
\caption{Agent B dialogue interaction (Danieli and Gerbino, 1995)}
\end{center}
\end{figure}
}

{\scriptsize
\begin{table}[ht]
\begin{center}
\begin{tabular}{|l|l|} \hline
attribute		& actual value	\\ \hline
depart-city		&Torino\\
arrival-city		&Milano\\
depart-range		&evening		\\
depart-time		&8pm		\\ \hline
\end{tabular}
\label{key}
\caption{Attribute value matrix instantiation, scenario key for Dialogues 1 and 2} 
\end{center}
\end{table}
}

Performance evaluation for an agent requires a corpus of dialogues
between users and the agent, in which users execute a set of
scenarios.  Each scenario execution has a corresponding AVM
instantiation indicating the task information requirements for the
scenario, where each attribute is paired with the attribute value
obtained via the dialogue. 

For example, assume that a scenario requires the user to find a train
from Torino to Milano that leaves in the evening, as in the longer
versions of Dialogues 1 and 2 in Figures 2 and 3.\footnote{These
dialogues have been slightly modified from~\cite{DanieliGerbino95}.
The  attribute names at the end of each utterance will be explained below.}
Table 2 contains an AVM corresponding to a ``key'' for this scenario.
All dialogues resulting from execution of this scenario in which the
agent and the user correctly convey all attribute values (as in
Figures 2 and 3) would have the same AVM as the scenario key in Table
2.  The AVMs of the remaining dialogues would differ from the key by
at least one value.  Thus, even though the dialogue strategies in
Figures 2 and 3 are radically different, the AVM
task representation for these dialogues is identical and the
performance of the system for the same task can thus be assessed on
the basis of the AVM representation.

\subsection{Measuring Task Success}
\label{succ-sec}

{\scriptsize
\begin{table*}[t]
\begin{center}
\begin{tabular}{|r | r r r r | r r r r | r r | r r r r |} \hline
& \multicolumn{14}{c|}{\it KEY} \\ \hline
&\multicolumn{4}{c|}{DEPART-CITY}
& \multicolumn{4}{c|}{ARRIVAL-CITY}
& \multicolumn{2}{c|}{DEPART-RANGE}
& \multicolumn{4}{c|}{DEPART-TIME} \\ \hline
{\it DATA}&v1 	&v2 	&v3 	&v4	&v5 	&v6	&v7 	&v8	&v9
	&v10 	&v11 	&v12 	&v13	&v14 	\\ \hline
v1	&{\bf22}&	&1	&	&3	&	&	&	&
&	&	&	&	&	\\
v2	&	&{\bf29}&	&	&	&	&	&	&
&	&	&	&	&	\\
v3	&4	&	&{\bf16}&4	&	&	&1	&	&
&	&	&	&	&	\\
v4	&1	&1	&5	&{\bf11}&	&	&1	&	&
&	&	&	&	&	\\ \hline
v5	&3	&	&	&	&{\bf20}&	&	&	&
&	&	&	&	&	\\
v6      &	&	&	&	&	&{\bf22}&	&	&
&	&	&	&	&	\\
v7	&	&	&2	&	&1	&1	&{\bf20}&5	&
&	&	&	&	&	\\
v8	&	&	&1	&	&1	&2	&8	&{\bf15}&
&	&	&	&	&	\\ \hline
v9	&	&	&	&	&	&	&	&
&{\bf45}&10	&	&	&	&	\\
v10	&	&	&	&	&	&	&	&	&5
     &{\bf40}&	&	&	&	\\ \hline
v11	&	&	&	&	&	&	&	&	&
&	&{\bf20}&	&2	&	\\
v12	&	&	&	&	&	&	&	&	&
&	&1	&{\bf19}&2	&4	\\
v13	&	&	&	&	&	&	&	&	&
&	&2	&	&{\bf18}&	\\
v14	&	&	&	&	&	&	&	&	&
&	&2	&6	&3	&{\bf21}\\ \hline
sum	&30 	&30 	&25 	&15 	&25 	&25 	&30 	&20 	&50
	&50 	&25 	&25 	&25 	&25	\\ \hline
\end{tabular}
\label{aclFM1}
\caption{Confusion matrix, Agent A}
\end{center}
\end{table*}
}

{\scriptsize
\begin{table*}[t]
\begin{center}
\begin{tabular}{|r | r r r r | r r r r | r r | r r r r |} \hline
	& \multicolumn{14}{c|}{\it KEY} \\ \hline
	&\multicolumn{4}{c|}{DEPART-CITY}
	& \multicolumn{4}{c|}{ARRIVAL-CITY}
	& \multicolumn{2}{c|}{DEPART-RANGE}
	& \multicolumn{4}{c|}{DEPART-TIME} \\ \hline
{\it DATA}&v1 	&v2 	&v3 	&v4	&v5 	&v6	&v7 	&v8	&v9
	&v10 	&v11 	&v12 	&v13	&v14 	\\ \hline
v1	&{\bf16}&	&1	&	&4	&	&	&	&3
&2	&	&	&	&\\
v2	&1	&{\bf20}&1	&	&	&3	&	&	&
&	&	&	&	&\\
v3	&5	&1	&{\bf9}&4	&2	&	&4	&2	&
&	&	&	&	&\\
v4	&1	&2	&6	&{\bf6}&	&	&2	&3	&
&	&	&	&	&\\ \hline
v5	&4	&	&	&	&{\bf15}&	&	&	&2
&3	&	&	&	&\\
v6      &1	&6	&	&	&	&{\bf19}&	&	&
&	&	&	&	&\\
v7	&	&	&5	&2	&1	&1	&{\bf15}&4	&
&	&	&	&	&\\
v8	&	&1	&3	&3	&1	&2	&9	&{\bf11}&
&	&	&	&	&\\ \hline
v9	&2	&	&	&	&2 	&	&	&
&{\bf39}&10	&	&	&	&\\
v10	&	&	&	&	&	&	&	&	&6
     &{\bf35}&	&	&	&\\ \hline
v11	&	&	&	&	&	&	&	&	&
&	&{\bf20}&5	&5	&4\\
v12	&	&	&	&	&	&	&	&	&
&	&	&{\bf10}&5	&5\\
v13	&	&	&	&	&	&	&	&	&
&	&5	&5	&{\bf10}&5\\
v14	&	&	&	&	&	&	&	&	&
&	&	&5	&5	&{\bf11}\\ \hline
sum	&30 	&30 	&25 	&15 	&25 	&25 	&30 	&20 	&50
	&50 	&25 	&25 	&25 	&25\\ \hline
\end{tabular}
\label{aclFM2}
\caption{Confusion matrix, Agent B}
\end{center}
\end{table*}
}

Success at the task for a whole dialogue (or subdialogue) is measured
by how well the agent and user achieve the information requirements of
the task by the end of the dialogue (or subdialogue). This section
explains how PARADISE uses the Kappa coefficient~\cite{Carletta96,Siegel88}
to operationalize the task-based success measure in Figure~\ref{objectives-fig}.

The Kappa coefficient, $\kappa$, is calculated from a confusion matrix
that summarizes how well an agent achieves the information
requirements of a particular task for a set of dialogues instantiating
a set of scenarios.\footnote{ Confusion matrices can be constructed to
summarize the result of dialogues for any subset of the scenarios,
attributes, users or dialogues.} For example, Tables 3 and 4 show two
hypothetical confusion matrices that could have been generated in an
evaluation of 100 complete dialogues with each of two train timetable
agents A and B (perhaps using the confirmation strategies illustrated
in Figures 2 and 3, respectively).\footnote{The distributions in the
tables were roughly based on performance results
in~\cite{DanieliGerbino95}.} The values in the matrix cells are based
on comparisons between the dialogue and scenario key AVMs.  Whenever
an attribute value in a dialogue (i.e., data) AVM {\it matches} the
value in its scenario key, the number in the appropriate diagonal cell
of the matrix (boldface for clarity) is incremented by 1.  The off
diagonal cells represent {\it misunderstandings} that are not
corrected in the dialogue.  Note that depending on the strategy that
a spoken dialogue agent uses, confusions across attributes are possible,
e.g., ``Milano '' could be confused with ``morning.''
The effect of misunderstandings that {\it
are} corrected during the course of the dialogue are reflected in
the costs associated with the dialogue, as will be discussed below.

The first matrix summarizes how the 100 AVMs representing each
dialogue with Agent A compare with the AVMs representing the relevant
scenario keys, while the second matrix summarizes the information
exchange with Agent B.  Labels v1 to v4 in each matrix represent the
possible values of depart-city shown in Table 1; v5 to v8 are for
arrival-city, etc. Columns represent the key, specifying which
information values the agent and user were supposed to communicate to
one another given a particular scenario.  (The equivalent column sums
in both tables reflects that users of both agents were assumed to have
performed the same scenarios). Rows represent the data collected from
the dialogue corpus, reflecting what attribute values were actually
communicated between the agent and the user.

Given a confusion matrix M,  success at achieving the
information requirements of the task is   measured with  the Kappa
coefficient~\cite{Carletta96,Siegel88}:
\[\kappa =  \frac {P(A) - P(E)}{1 - P(E)} \]
P(A) is the proportion of times that the AVMs for the actual set of
dialogues agree with the AVMs for the scenario keys, and P(E) is the
proportion of times that the AVMs for the dialogues and the keys are
expected to agree by chance.\footnote{$\kappa$ has been used to
measure pairwise agreement among coders making category
judgments~\cite{Carletta96,Krippendorf80,Siegel88}. Thus, the observed
user/agent interactions are modeled as a coder, and the ideal
interactions as an expert coder.}  When there is no agreement other
than that which would be expected by chance, $\kappa$ $=$ 0.  When
there is total agreement, $\kappa$ $=$ 1. $\kappa$ is superior to
other measures of success such as transaction
success~\cite{DanieliGerbino95}, concept
accuracy~\cite{SimpsonFraser93}, and percent agreement~\cite{GCY}
because $\kappa$ takes into account the inherent complexity of the
task by correcting for chance expected agreement.  Thus $\kappa$
provides a basis for comparisons across agents that are performing
{\it different} tasks.

When the prior distribution of the categories is unknown,
P(E), the expected chance agreement between the data and the key, can be
estimated from the distribution of the values  in the keys. This can be 
calculated from  confusion matrix M, since the columns represent the
values in the keys.  In particular:
\[P(E) = \sum_{i=1}^{n} (\frac{t_{i}}{T})^2\]
where $t_{i}$ is the sum of the frequencies in column i of M, and $T$
is the sum of the frequencies in M ($t_{1}$ + \ldots + $t_{n}$).

P(A), the actual agreement
between the data and the key, is always
computed from the confusion matrix M:
\[P(A) = \frac{\sum_{i=1}^{n} M(i,i)}{T} \]

Given the confusion matrices in Tables 3 and 4, P(E) = 0.079 for both
agents.\footnote{Using a single confusion matrix for all attributes as
in Tables 3 and 4 inflates $\kappa$ when there are few cross-attribute
confusions by making P(E) smaller.
In some cases it might be desirable to calculate
$\kappa$ first for identification of attributes and then for values
within attributes, or to average $\kappa$ for each attribute to
produce an overall $\kappa$ for the task.}  For Agent A, P(A) = 0.795
and $\kappa$ = 0.777, while for Agent B, P(A) = 0.59 and $\kappa$ =
0.555, suggesting that Agent A is more successful than B in achieving
the task goals.

\subsection{Measuring Dialogue Costs}
\label{cost-sec}

As shown in Figure \ref{objectives-fig}, performance is also a
function of a combination of cost measures.  Intuitively, cost
measures should be calculated on the basis of any user or agent
dialogue behaviors that should be minimized. A wide range of cost
measures have been used in previous work; these include pure
efficiency measures such as the number of turns or elapsed time to
complete the task~\cite{AbellaEtAl96,HirschmanEtAl90,SmithGordon97,Walker96a}, as well as
measures of qualitative phenomena such as inappropriate or repair
utterances~\cite{DanieliGerbino95,HirschmanPao93,SimpsonFraser93}.

PARADISE represents each cost measure as a function $c_i$ that can be
applied to any (sub)dialogue. First, consider the simplest case of
calculating efficiency measures over a whole dialogue.  For example,
let $c_{1}$ be the total number of utterances.  For the whole dialogue
D1 in Figure 2, $c_{1}$(D1) is 23 utterances.  For the whole dialogue
D2 in Figure 3, $c_{1}$(D2) is 10 utterances.

To calculate costs over subdialogues and for some of the qualitative measures,
it is necessary to be able to specify which
information goals each utterance contributes to.  PARADISE uses its AVM
representation to link the information goals of the task
to any arbitrary dialogue behavior, by tagging the dialogue with the
attributes for the task.\footnote{This tagging can be hand generated,
or system generated and hand corrected.  
Preliminary studies indicate that reliability for human tagging is higher
for AVM attribute tagging than for other types of discourse segment
tagging~\cite{PassonneauLitman97,HirschbergNakatani96}.}
This makes it possible to evaluate any potential dialogue strategies
for achieving the task, as well as to evaluate dialogue strategies that
operate at the level of dialogue subtasks (subdialogues).

\begin{figure}[htb]
\centerline{\psfig{figure=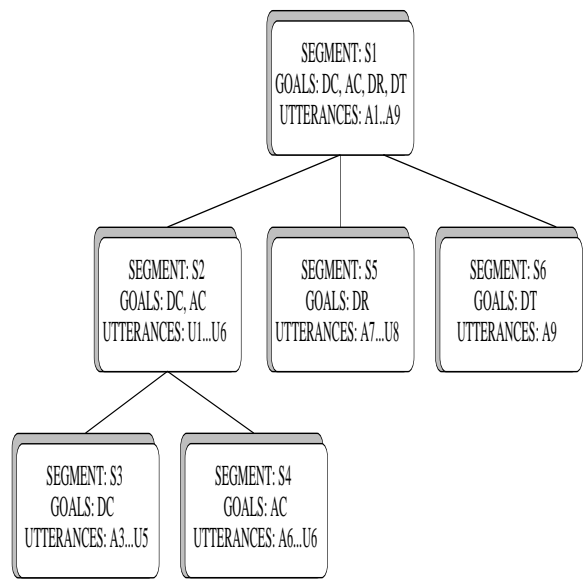,height=3.0in,width=3.0in}}
\caption{Task-defined discourse structure of Agent A dialogue interaction}
\end{figure}

Consider the longer versions of Dialogues 1 and 2 in
Figures 2 and 3.  Each utterance in Figures 2 and 3 has been tagged
using one or more of the attribute abbreviations in Table 1, according
to the subtask(s) the utterance contributes to. As a convention of
this type of tagging, utterances that contribute to the success of the
whole dialogue, such as greetings, are tagged with all the attributes.
Since the structure of a dialogue reflects the structure of the task~\cite{Carberry89,GS86,LA90}, the tagging of a dialogue by the AVM
attributes can be used to generate a hierarchical discourse structure
such as that shown in Figure 4 for Dialogue 1 (Figure 2).
For example, segment (subdialogue) S2 in Figure 4 is about
both depart-city (DC) and arrival-city (AC). It contains segments S3
and S4 within it, and consists of utterances U1 $\ldots$ U6.

Tagging by AVM attributes is required to calculate costs
over subdialogues, since for any subdialogue, task attributes define
the subdialogue. For subdialogue S4 in Figure 4, which is about the
attribute arrival-city and consists of utterances A6 and U6,
$c_{1}$(S4) is 2.

Tagging by AVM attributes is also required to calculate the cost of some of
the qualitative measures, such as number of repair utterances. (Note
that to calculate such costs, each utterance in the corpus of
dialogues must also be tagged with respect to the qualitative
phenomenon in question, e.g. whether the utterance is a
repair.\footnote{Previous work has shown that this can be done with
high reliability~\cite{HirschmanPao93}.})  For example, let $c_{2}$ be
the number of repair utterances.  The repair utterances in Figure 2
are A3 through U6, thus $c_{2}$(D1) is 10 utterances and $c_{2}$(S4) is 2
utterances. The repair
utterance in Figure 3 is U2, but note that according to the AVM task
tagging, U2 simultaneously addresses the information goals for
depart-range. In general, if an utterance U contributes to the
information goals of N different attributes, each attribute accounts
for 1/N of any costs derivable from U. Thus, $c_{2}$(D2) is .5.

Given a set of $c_i$, it is necessary to combine the different cost
measures in order to determine their relative contribution to
performance.  The next section explains how to combine $\kappa$ with a
set of $c_i$ to yield an overall performance measure.


\subsection{Estimating a Performance Function}
\label{perf-func-sec}

Given the definition of success and costs above and the model in
Figure~\ref{objectives-fig}, performance for any (sub)dialogue D is
defined as follows:\footnote{We assume an additive performance
(utility) function because it appears that $\kappa$ and the various
cost factors $c_i$ are utility independent and additive
independent~\cite{KeeneyRaiffa76}.
It is possible however that user satisfaction data collected in future
experiments (or other data such as willingness to pay or use) would
indicate otherwise. If so, continuing use of an additive
function might require a transformation of the data, a reworking of
the model shown in Figure \ref{objectives-fig}, or the inclusion of
interaction terms in the model~\cite{Cohen95}.}

\[{\rm Performance} = (\alpha \ast {\cal N}(\kappa)) -  \sum_{i=1}^{n} w_{i}
\ast {\cal N}(c_{i})\]
Here $\alpha$ is a weight on $\kappa$, the cost functions
$c_i$ are weighted by $w_i$, and {$\cal N$} is a Z score normalization
function~\cite{Cohen95}.  

The normalization function is used to overcome the problem that the
values of $c_i$ are not on the same scale as $\kappa$, and that the
cost measures $c_i$ may also be calculated over widely varying scales
(e.g. response delay could be measured using seconds while, in the
example, costs were calculated in terms of number of utterances). This
problem is easily solved by normalizing each factor {\it x} to its Z score:
\[{\cal N}(x) = \frac{x - \overline{x}}{\sigma_{{\it x}}}\]
where $\sigma_{x}$ is the standard deviation for {\it x}.  

{\scriptsize
\begin{table}[htb]
\begin{center}
\begin{tabular}{|r|r|r|l|r|r|} \hline
user &agent  &US     &$\kappa$	&$c_1$ (\#utt)	&$c_2$ (\#rep)\\ \hline
1       &A      &1      &1       	&46      &30 \\
2       &A      &2      &1       	&50      &30\\
3       &A      &2      &1       	&52      &30\\
4       &A      &3      &1       	&40      &20\\
5       &A      &4      &1       	&23      &10\\
6       &A      &2      &1       	&50      &36\\
7       &A      &1      &0.46   	&75      &30\\
8       &A      &1      &0.19   	&60      &30\\ \hline
9       &B      &6      &1       	&8       &0\\
10      &B      &5      &1       	&15      &1\\
11      &B      &6      &1       	&10      &0.5\\
12      &B      &5      &1       	&20      &3\\
13      &B      &1      &0.19   	&45      &18\\
14      &B      &1      &0.46   	&50      &22\\
15      &B      &2      &0.19   	&34      &18\\
16      &B      &2      &0.46   	&40      &18\\ \hline
Mean(A)  &A      &2      &0.83   	&49.5    &27\\
Mean(B)  &B      &3.5      &0.66   	&27.8    &10.1\\
Mean    &NA      &2.75      &0.75   	&38.6      &18.5\\ \hline
\end{tabular}
\label{perf}
\caption{Hypothetical performance data from users of Agents A and B}
\end{center}
\end{table} }

To illustrate the method for estimating a performance function, we
will use a subset of the data from Tables 3 and 4, shown in Table 5.
Table 5 represents the results from a hypothetical experiment in which
eight users were randomly assigned to communicate with Agent A and eight users were
randomly assigned to communicate with Agent
B. Table 5 shows user satisfaction (US) ratings (discussed below),
$\kappa$, number of utterances (\#utt) and number of repair utterances
(\#rep) for each of these users.  Users 5 and 11 correspond to the
dialogues in Figures 2 and 3 respectively.  To normalize $c_{1}$ for
user 5, we determine that $\overline{c_{1}}$ is 38.6 and
$\sigma_{c_1}$ is 18.9. Thus, ${\cal N}(c_{1})$ is -0.83. Similarly
${\cal N}(c_{1})$ for user 11 is -1.51.

To estimate the performance function, the weights $\alpha$ and $w_i$
must be solved for.  Recall that the claim implicit in
Figure~\ref{objectives-fig} was that the relative contribution of task
success and  dialogue costs to performance should be
calculated by considering their contribution to user satisfaction.
User satisfaction is typically calculated with surveys that ask users
to specify the degree to which they agree with one or more statements
about the behavior or the performance of the system. A single user
satisfaction measure can be calculated from a single question, or as
the mean of a set of ratings.  The hypothetical user satisfaction
ratings shown in Table 5 range from a high of 6 to a low of 1.
 
Given a set of dialogues for which user satisfaction (US), $\kappa$
and the set of $c_i$ have been collected experimentally, the weights
$\alpha$ and $w_i$ can be solved for using multiple linear regression.
Multiple linear regression produces a set of coefficients (weights)
describing the relative contribution of each predictor factor in
accounting for the variance in a predicted factor. In this case, on
the basis of the model in Figure \ref{objectives-fig}, US is treated
as the predicted factor. Normalization of the predictor factors
($\kappa$ and $c_i$) to their Z scores guarantees that the relative
magnitude of the coefficients directly indicates the relative
contribution of each factor.  Regression on the Table 5 data for both
sets of users tests which factors $\kappa$, \#utt, \#rep most strongly
predicts US.

In this illustrative example, the results of the regression with all
factors included shows that only $\kappa$ and \#rep are significant (p
$<$ .02). In order to develop a performance function estimate that includes
only significant factors and eliminates redundancies,
a second regression including only significant factors must then
be done. In this case, a second regression yields the predictive
equation:

\[{\rm Performance} = .40 {\cal N}(\kappa) - .78{\cal N}(c_2)  \]
\noindent
i.e., $\alpha$ is .40 and $w_2$ is .78.  The results also show
$\kappa$ is significant at p $<$ .0003, \#rep significant at p $<$
.0001, and the combination of $\kappa$ and \#rep account for 92\% of
the variance in US, the external validation criterion.  The factor
\#utt was not a significant predictor of performance, in part because
\#utt and \#rep are highly redundant. (The correlation between \#utt and
\#rep is 0.91).

Given these predictions about the relative contribution of different
factors to performance, it is then possible to return to the problem
first introduced in Section \ref{intro-sec}: given potentially
conflicting performance criteria such as robustness and efficiency,
how can the performance of Agent A and Agent B be compared?  Given
values for $\alpha$ and $w_i$, performance can be calculated for both
agents using the equation above. The mean performance of A is -.44 and
the mean performance of B is .44, suggesting that Agent B may perform
better than Agent A overall.

The evaluator must then however test these performance differences for
statistical significance. In this case, a {\it t} test shows that
differences are only significant at the p $<$~.07 level, indicating a
trend only. In this case, an evaluation over a larger subset of the
user population would probably show significant differences.

\subsection{Application to Subdialogues}
\label{subdialogues-sec}

Since both $\kappa$ and $c_i$ can be calculated over subdialogues,
performance can also be calculated at the subdialogue level by using
the values for $\alpha$ and $w_i$ as solved for above. This
assumes that the factors that are predictive of global performance,
based on US, generalize as predictors of local performance,
i.e. within subdialogues defined by subtasks, as defined by the 
attribute tagging.\footnote{This assumption has a sound basis in theories of
dialogue structure~\cite{Carberry89,GS86,LA90}, but should be tested
empirically.}

Consider calculating the performance of the dialogue strategies used by train
timetable Agents A and B, over the subdialogues that repair the value of
depart-city.  Segment S3 (Figure 4) is an example of such a
subdialogue with Agent A.
As in the initial estimation of a performance function, our analysis
requires experimental data, namely a set of values for $\kappa$ and
$c_i$, and the application of the Z score normalization function to
this data.  However, the values for $\kappa$ and $c_i$ are now
calculated at the subdialogue rather than the whole dialogue level.
In addition, only data from comparable strategies can be used to
calculate the mean and standard deviation for normalization.
Informally, a comparable strategy is one which applies in the same
state and has the same effects.


For example, to calculate $\kappa$ for Agent A over the
subdialogues that repair depart-city, P(A) and P(E) are computed using
only the subpart of Table 3 
concerned with depart-city.  For Agent A, P(A) = .78, P(E) = .265, and
$\kappa$ = .70. Then, this value of $\kappa$ is normalized using
data from comparable subdialogues with both
Agent A and Agent B.  Based on the data in Tables 3 and 4, the mean $\kappa$ is .515 and $\sigma$ is .261,
so that ${\cal N}(\kappa)$ for Agent A is .71.

To calculate $c_2$ for Agent A,
assume
that the average number of repair utterances for Agent A's subdialogues
that repair depart-city is 6, that the mean over all comparable repair subdialogues is 4, and
the standard deviation is 2.79.  Then ${\cal N}(c_2)$ is .72.


Let Agent A's repair dialogue strategy for subdialogues repairing depart-city 
be R$_A$ and Agent B's repair strategy for depart-city be R$_B$. Then
using the performance equation above, predicted performance for R$_A$
is:

\[{\rm Performance(R_A)} = .40 \ast .71 - .78 \ast .72 = -0.28  \]

For Agent B, using the appropriate subpart of Table 4 to calculate $\kappa$,
assuming that the average number of depart-city repair utterances
is 1.38, and using similar calculations, yields
\[{\rm Performance(R_B)} = .40 \ast -.71 - .78 \ast -.94 =  0.45  \]

Thus the results of these experiments predict that when an agent needs
to choose between the repair strategy that Agent B uses and the repair
strategy that Agent A uses for repairing depart-city, it should use Agent B's strategy R$_B$,
since the performance(R$_B$) is predicted to be greater than the
performance(R$_A$).


Note that the ability to calculate performance over subdialogues
allows us to conduct experiments that simultaneously test multiple
dialogue strategies.  For example, suppose Agents A and B had
different strategies for presenting the value of depart-time (in
addition to different confirmation strategies).  Without the ability
to calculate performance over subdialogues, it would be impossible to
test the effect of the different presentation strategies independently
of the different confirmation strategies.

\subsection{Summary}
\label{summary-sec}

We have presented the PARADISE framework, and have used it to
evaluate two hypothetical dialogue agents in a simplified
train timetable task domain.  We used PARADISE to derive a
performance function for this task, by
estimating the relative contribution of a set of potential predictors
to user satisfaction.  The PARADISE methodology consists of the
following steps:
\begin{itemize}
\item definition of a task and a set of scenarios;
\item specification of the AVM task representation;
\item experiments with alternate dialogue agents for the task;
\item calculation of user satisfaction using surveys;
\item calculation of task success using $\kappa$;
\item calculation of dialogue cost using efficiency and qualitative measures;
\item estimation of a performance function using linear regression and values for user satisfaction, $\kappa$ and dialogue costs;
\item comparison with other agents/tasks to determine which factors generalize;
\item refinement of the performance model.
\end{itemize}

Note that all of these steps are required to develop the performance
function. However once the weights in the performance function have
been solved for, user satisfaction ratings no longer need to be
collected.  Instead, predictions about user satisfaction can be made
on the basis of the predictor variables, as illustrated in the
application of PARADISE to subdialogues.

Given the current state of knowledge, it is important to emphasize
that researchers should be cautious about generalizing a derived performance
function to other agents or tasks.
Performance function estimation should be done
iteratively over many different tasks
and dialogue strategies to see which factors generalize.  In this way, the field can make progress on
identifying the relationship between various factors and can move
towards more predictive models of spoken dialogue agent performance.

\section{Generality}
\label{gen-sec}

In the previous section we used PARADISE to 
evaluate two confirmation strategies, using as examples fairly
simple information access dialogues in the train timetable domain.
In this section we
demonstrate that PARADISE is applicable to a range of tasks, domains, and
dialogues, by presenting AVMs for two tasks  involving more than
information access, and showing how additional dialogue
phenomena can be tagged using AVM attributes.

{ \scriptsize
\begin{table}[htb]
\begin{center}
\begin{tabular}{|l|l|c|} \hline
attribute  &possible values & information flow\\ \hline
depart-city (DC) &Milano, Roma, Torino, Trento & to agent \\
arrival-city (AC) &Milano, Roma, Torino, Trento & to agent \\
depart-range (DR) &morning,evening & to agent\\
depart-time (DT) &6am,8am,6pm,8pm & to user\\ 
request-type (RT) & reserve, purchase & to agent \\ \hline
\end{tabular}
\label{tt-req-AVM}
\caption{Attribute value matrix, train timetable domain with requests}
\end{center}
\end{table} }

First, consider an extension of the train timetable task, where
an agent can handle requests to reserve a seat or purchase a
ticket.  This task could be represented using the AVM in Table 6
(an extension of Table 1), where the
agent must now acquire the value of the attribute request-type,
in order to know what to do with the other information it has
acquired.

{\scriptsize
\begin{figure}[ht]
\begin{center}
\begin{tabular}{rll}
	U1: &	I want to go from Torino to Roma   &DC,AC \\
	C1: & 	Approximately what time of day would you like to travel? & DR \\
	U2: &	What are the options?  & DR \\
	C2: &	Morning or evening. & DR \\
	U3: & 	Are those departure times?  &DR \\
	C3: &	Yes.  &DR \\
	U4: &	I'd like to leave in the morning. & DR \\
	C4: &	Train 702 leaves Torino Porto at 8 a.m. & DT \\
	U5: &	Please reserve me a seat on that train. & RT 
\end{tabular}
\label{d3}
\caption{Hypothetical Agent C dialogue interaction}
\end{center}
\end{figure}
}

Figure 5 presents a hypothetical dialogue in this extended task
domain, and illustrates user utterance types and an agent dialogue strategy
that are very different from those in Figures 2 and 3.  First, Agent C in Figure
5 uses a ``no confirmation'' dialogue strategy, in contrast to
the explicit and implicit confirmation strategies used in Figures 2
and 3.  Second, Figure 5 illustrates new types of {\it user} utterances that do not
directly further the informational goals of the task. 
In U2, the
user asks the agent a wh-question about the DR attribute itself,
rather than providing information about that attribute's value. Since
U2 satisfies a knowledge precondition related to answering C1, U2
contributes to the DR goal and is tagged as such.  In U3, the user
similarly asks a yes-no question 
that addresses a subgoal related to
answering C1.  Finally, U5 illustrates a user request for
an agent action, and is tagged with the RT attribute. 
The value of RT in the AVM instantiation for the dialogue
would be ``reserve.''

Second, consider the very different domain and task of diagnosing a
fault and repairing a circuit~\cite{SmithGordon97}.  Figure 6 presents
one dialogue from this domain.

{\scriptsize
\begin{figure}[ht]
\begin{center}
\begin{tabular}{rll}
 1. C: & This is the circuit fix it shop. & ID,CB,RB,FT,FC,T\\
	& How may I help you? & ID,CB,RB,FT,FC,T\\
 2. U: & I want to fix a circuit. & ID,CB,RB,FT,FC,T\\
 3. C: & What is the ID of the circuit? & ID\\
 4. U: & Rs111. & ID\\
 5. C: & I am familiar with that circuit. & CB\\
	& The LED is supposed to be displaying&\\
       &  alternately flashing one and seven. & CB\\
 6. U: & The LED is off. &RB\\
 7. C: & What is the switch at when the LED is off? &RB\\
 8. U: & The switch is down. &RB\\
 9. C: & Put the switch up. &RB\\
10. U: & Okay. &RB\\
11. C: & Put the knob to one zero. &RB\\
12. U: & Okay. &RB\\
13. C: & What is the LED displaying? &RB\\
14. U: & LED is displaying only a flashing seven.&RB\\
15. C: & Is there a wire between connector eight four  &\\
	& and connector nine nine? &FT\\
16. U: & No. &FT\\
17. C: & Add a wire between connector eight four & \\
       &  and connector nine nine. &FC\\
18. U: & Done. &FC\\
19. C: & What is the LED displaying? &T\\
20. U: & Alternately flashing one and seven. &T\\
21. C: & Is the one on the LED displaying for & \\
	&  a longer period of time? &T\\
22. U: & No. &T\\
23. C: & Is the seven on the LED displaying & \\
	& for a longer period of time? &T\\
24. U: & No. &T\\
25. C: & Put the knob to zero. &T\\
26. U: & Okay. &T\\\
27. C: & What is the LED displaying? &T\\
28. U: & Alternately displaying one and seven. &T\\
29. C: & Is the one on the LED displaying & \\
	 & for a longer period of time? &T\\
30. U: & Yes. &T\\
31. C: & Put the switch down. &T\\
32. U: & The switch is down. &T\\
33. C: & What is the LED displaying? &T\\
34. U: & Nothing. &T\\
35. C: & The circuit is working correctly.  Good-bye. &T\\
\end{tabular}
\end{center}
\caption{A circuit domain dialogue (Smith and Gordon, 1997), with AVM tagging}
\end{figure}
}

Smith and Gordon collected 144 dialogues for this task, in which agent
initiative was varied by using different dialogue strategies, and
tagged each dialogue according to the following subtask
structure:\footnote{They report a $\kappa$ of .82 for reliability of
their tagging scheme.}
\begin{itemize}
\item Introduction  (I)---establish the purpose of the task
\item Assessment  (A)---establish the current behavior
\item Diagnosis  (D)---establish the cause for the errant behavior
\item Repair  (R)---establish that the correction for the errant behavior has been made
\item Test  (T)---establish that the behavior is now correct
\end{itemize}
Our informational analysis of this task results in the AVM shown in
Table 7.  Note that the attributes are almost identical to Smith and
Gordon's list of subtasks.  Circuit-ID corresponds to Introduction,
Correct-Circuit-Behavior and Current-Circuit-Behavior correspond to
Assessment, Fault-Type corresponds to Diagnosis, Fault-Correction
corresponds to Repair, and Test corresponds to Test.  The attribute names
emphasize information exchange, while the subtask names emphasize 
function.
{ \scriptsize
\begin{table}[htb]
\begin{center}
\begin{tabular}{|l|l|} \hline
attribute  &possible values \\ \hline
Circuit-ID (ID)& RS111, RS112, ...  \\
Correct-Circuit-Behavior (CB) & Flash-1-7, Flash-1, ...  \\
Current-Circuit-Behavior (RB) & Flash-7 \\
Fault-Type (FT) & MissingWire84-99, MissingWire88-99, ... \\ 
Fault-Correction (FC) & yes, no \\ 
Test (T)& yes, no  \\ \hline
\end{tabular}
\label{CI-AVM}
\caption{Attribute value matrix,  circuit domain}
\end{center}
\end{table} }

Figure 6 is tagged with the
attributes from Table 7.  Smith and Gordon's
tagging of this dialogue according to their subtask representation was as follows: turns 1-4 were I, turns 5-14 were A, turns 15-16 were D,
turns 17-18 were R, and turns 19-35 were T.  Note that 
there are only two  differences between the dialogue
structures yielded by the two tagging schemes.
First, in our scheme (Figure 6), the greetings (turns 1 and 2) are tagged with all the attributes.
Second, Smith and Gordon's single tag A corresponds to two 
attribute tags in Table 7, which in our scheme
defines an extra level of structure within assessment
subdialogues.

\section {Discussion}
\label{conc-sec}


This paper presented the PARADISE framework for evaluating spoken
dialogue agents. PARADISE is a general framework for evaluating spoken
dialogue agents that integrates and enhances previous work.  PARADISE
supports comparisons among dialogue strategies with a task
representation that decouples {\it what} an agent needs to achieve in
terms of the task requirements from {\it how} the agent carries out
the task via dialogue. Furthermore, this task representation supports
the calculation of performance over subdialogues as well as whole
dialogues.  In addition, because PARADISE's success measure normalizes
for task complexity, it provides a basis for comparing agents
performing {\it different} tasks.

The PARADISE performance measure is a function of both task success
($\kappa$) and dialogue costs ($c_i$), and has a number of advantages.
First, it allows us to evaluate performance at any level of a
dialogue, since $\kappa$ and $c_i$ can be calculated for any dialogue
subtask.  Since performance can be measured over any subtask, and
since dialogue strategies can range over subdialogues or the whole
dialogue, we can associate performance with individual dialogue
strategies. Second, because our success measure $\kappa$ takes into
account the complexity of the task, comparisons can be made across
dialogue tasks.  Third, $\kappa$ allows us to measure partial success
at achieving the task.  Fourth, performance can combine both objective
and subjective cost measures, and specifies how to evaluate the
relative contributions of those costs factors to overall performance.
Finally, to our knowledge, we are the first to propose using user
satisfaction to determine weights on factors related to performance.

In addition, this approach is broadly integrative, incorporating
aspects of transaction success, concept accuracy, multiple cost
measures, and user satisfaction.  In our framework, transaction
success is reflected in $\kappa$, corresponding to dialogues with a
P(A) of 1.  Our performance measure also captures information similar
to concept accuracy, where low concept accuracy scores translate into
either higher costs for acquiring information from the user, or lower
$\kappa$ scores.

One limitation of the PARADISE approach is that the task-based success
measure does not reflect that some solutions might be better than
others.  For example, in the train timetable domain, we might like our
task-based success measure to give higher ratings to agents that
suggest express over local trains, or that provide helpful information
that was not explicitly requested, especially since the better
solutions might occur in dialogues with higher costs. It might be
possible to address this limitation by using the interval scaled data
version of $\kappa$~\cite{Krippendorf80}.  Another possibility is to
simply substitute a domain-specific task-based success measure in the
performance model for $\kappa$.

The evaluation model presented here has many applications in apoken
dialogue processing. We believe that the framework is also applicable
to other dialogue modalities, and to human-human task-oriented
dialogues.  In addition, while there are many proposals in the
literature for algorithms for dialogue strategies that are
cooperative, collaborative or helpful to the user
\cite{WJ82,PHW82,JWW84,ChuCarberry95}, very few of these strategies
have been evaluated as to whether they improve any measurable aspect
of a dialogue interaction. As we have demonstrated here, any dialogue
strategy can be evaluated, so it should be possible to show that a
cooperative response, or other cooperative strategy, actually improves
task performance by reducing costs or increasing task success.  We
hope that this framework will be broadly applied in future dialogue
research.

\section{Acknowledgments}

We would like to thank James Allen, Jennifer Chu-Carroll, Morena Danieli, Wieland Eckert, Giuseppe
Di Fabbrizio, Don Hindle, Julia Hirschberg, Shri Narayanan, Jay
Wilpon, Steve Whittaker and three anonymous reviews for helpful
discussion and comments on earlier versions of this paper.


\begin{thebibliography}{}

\bibitem[\protect\citename{Abella, Brown, and Buntschuh}1996]{AbellaEtAl96}
Abella, Alicia, Michael~K Brown, and Bruce Buntschuh.
\newblock 1996.
\newblock Development principles for dialog-based interfaces.
\newblock In {\em ECAI-96 Spoken Dialog Processing Workshop}, Budapest,
  Hungary.

\bibitem[\protect\citename{Bates and Ayuso}1993]{BatesAyuso}
Bates, Madeleine and Damaris Ayuso.
\newblock 1993.
\newblock A proposal for incremental dialogue evaluation.
\newblock In {\em Proceedings of the DARPA Speech and NL Workshop}, pages
  319--322.

\bibitem[\protect\citename{Carberry}1989]{Carberry89}
Carberry, S.
\newblock 1989.
\newblock Plan recognition and its use in understanding dialogue.
\newblock In A.~Kobsa and W.~Wahlster, editors, {\em User Models in Dialogue
  Systems}. Springer Verlag, Berlin, pages 133--162.

\bibitem[\protect\citename{Carletta}1996]{Carletta96}
Carletta, Jean~C.
\newblock 1996.
\newblock Assessing the reliability of subjective codings.
\newblock {\em Computational Linguistics}, 22(2):249--254.

\bibitem[\protect\citename{Chu-Carrol and Carberry}1995]{ChuCarberry95}
Chu-Carrol, Jennifer and Sandra Carberry.
\newblock 1995.
\newblock Response generation in collaborative negotiation.
\newblock In {\em Proceedings of the Conference of the 33rd Annual Meeting of
  the Association for Computational Linguistics}, pages 136--143.

\bibitem[\protect\citename{Cohen}1995]{Cohen95}
Cohen, Paul.~R.
\newblock 1995.
\newblock {\em Empirical Methods for Artificial Intelligence}.
\newblock MIT Press, Boston.

\bibitem[\protect\citename{Danieli \bgroup et al.\egroup }1992]{DanieliEtAl92}
Danieli, M., W.~Eckert, N.~Fraser, N.~Gilbert, M.~Guyomard, P.~Heisterkamp,
  M.~Kharoune, J.~Magadur, S.~McGlashan, D.~Sadek, J.~Siroux, and N.~Youd.
\newblock 1992.
\newblock Dialogue manager design evaluation.
\newblock Technical Report Project Esprit 2218 SUNDIAL, WP6000-D3.

\bibitem[\protect\citename{Danieli and Gerbino}1995]{DanieliGerbino95}
Danieli, Morena and Elisabetta Gerbino.
\newblock 1995.
\newblock Metrics for evaluating dialogue strategies in a spoken language
  system.
\newblock In {\em Proceedings of the 1995 AAAI Spring Symposium on Empirical
  Methods in Discourse Interpretation and Generation}, pages 34--39.

\bibitem[\protect\citename{Doyle}1992]{Doyle92}
Doyle, Jon.
\newblock 1992.
\newblock Rationality and its roles in reasoning.
\newblock {\em Computational Intelligence}, 8(2):376--409.

\bibitem[\protect\citename{Fraser}1995]{Fraser95}
Fraser, Norman~M.
\newblock 1995.
\newblock Quality standards for spoken dialogue systems: a report on progress
  in {EAGLES}.
\newblock In {\em ESCA Workshop on Spoken Dialogue Systems Vigso, Denmark},
  pages 157--160.

\bibitem[\protect\citename{Gale, Church, and Yarowsky}1992]{GCY}
Gale, William, Ken~W. Church, and David Yarowsky.
\newblock 1992.
\newblock Estimating upper and lower bounds on the performance of word-sense
  disambiguation programs.
\newblock In {\em Proc. of 30th ACL}, pages 249--256, Newark, Delaware.

\bibitem[\protect\citename{Grosz and Sidner}1986]{GS86}
Grosz, Barbara~J. and Candace~L. Sidner.
\newblock 1986.
\newblock Attentions, intentions and the structure of discourse.
\newblock {\em Computational Linguistics}, 12:175--204.

\bibitem[\protect\citename{Hirschberg and Nakatani}1996]{HirschbergNakatani96}
Hirschberg, Julia and Christine Nakatani.
\newblock 1996.
\newblock A prosodic analysis of discourse segments in direction-giving
  monologues.
\newblock In {\em 34th Annual Meeting of the Association for Computational
  Linguistics}, pages 286--293.

\bibitem[\protect\citename{Hirschman \bgroup et al.\egroup
  }1990]{HirschmanEtAl90}
Hirschman, Lynette, Deborah~A. Dahl, Donald~P. McKay, Lewis~M. Norton, and
  Marcia~C. Linebarger.
\newblock 1990.
\newblock Beyond class {A}: A proposal for automatic evaluation of discourse.
\newblock In {\em Proceedings of the Speech and Natural Language Workshop},
  pages 109--113.

\bibitem[\protect\citename{Hirschman and Pao}1993]{HirschmanPao93}
Hirschman, Lynette and Christine Pao.
\newblock 1993.
\newblock The cost of errors in a spoken language system.
\newblock In {\em Proceedings of the Third European Conference on Speech
  Communication and Technology}, pages 1419--1422.

\bibitem[\protect\citename{Joshi, Webber, and Weischedel}1984]{JWW84}
Joshi, Aravind~K., Bonnie~L. Webber, and Ralph~M. Weischedel.
\newblock 1984.
\newblock Preventing false inferences.
\newblock In {\em COLING84: Proc. 10th International Conference on
  Computational Linguistics.}, pages 134--138.

\bibitem[\protect\citename{Kamm}1995]{Kamm95}
Kamm, Candace.
\newblock 1995.
\newblock User interfaces for voice applications.
\newblock In David Roe and Jay Wilpon, editors, {\em Voice Communication
  between Humans and Machines}. National Academy Press, pages 422--442.

\bibitem[\protect\citename{Keeney and Raiffa}1976]{KeeneyRaiffa76}
Keeney, Ralph and Howard Raiffa.
\newblock 1976.
\newblock {\em Decisions with Multiple Objectives: Preferences and Value
  Tradeoffs}.
\newblock John Wiley and Sons.

\bibitem[\protect\citename{Krippendorf}1980]{Krippendorf80}
Krippendorf, Klaus.
\newblock 1980.
\newblock {\em Content Analysis: An Introduction to its Methodology}.
\newblock Sage Publications, Beverly Hills, Ca.

\bibitem[\protect\citename{Litman and Allen}1990]{LA90}
Litman, Diane and James Allen.
\newblock 1990.
\newblock Recognizing and relating discourse intentions and task-oriented
  plans.
\newblock In Philip Cohen, Jerry Morgan, and Martha Pollack, editors, {\em
  Intentions in Communication}. MIT Press.

\bibitem[\protect\citename{Passonneau and Litman}1997]{PassonneauLitman97}
Passonneau, Rebecca~J. and Diane Litman.
\newblock 1997.
\newblock Discourse segmentation by human and automated means.
\newblock {\em Computational Linguistics}, 23(1).

\bibitem[\protect\citename{Polifroni \bgroup et al.\egroup }1992]{PHSZ92}
Polifroni, Joseph, Lynette Hirschman, Stephanie Seneff, and Victor Zue.
\newblock 1992.
\newblock Experiments in evaluating interactive spoken language systems.
\newblock In {\em Proceedings of the DARPA Speech and NL Workshop}, pages
  28--33.

\bibitem[\protect\citename{Pollack, Hirschberg, and Webber}1982]{PHW82}
Pollack, Martha, Julia Hirschberg, and Bonnie Webber.
\newblock 1982.
\newblock User participation in the reasoning process of expert systems.
\newblock In {\em Proceedings First National Conference on Artificial
  Intelligence}, pages pp. 358--361.

\bibitem[\protect\citename{Shriberg, Wade, and Price}1992]{SWP92}
Shriberg, Elizabeth, Elizabeth Wade, and Patti Price.
\newblock 1992.
\newblock Human-machine problem solving using spoken language systems ({SLS}):
  Factors affecting performance and user satisfaction.
\newblock In {\em Proceedings of the DARPA Speech and NL Workshop}, pages
  49--54.

\bibitem[\protect\citename{Siegel and Castellan}1988]{Siegel88}
Siegel, Sidney and N.~J. Castellan.
\newblock 1988.
\newblock {\em Nonparametric Statistics for the Behavioral Sciences}.
\newblock McGraw Hill.

\bibitem[\protect\citename{Simpson and Fraser}1993]{SimpsonFraser93}
Simpson, A. and N.~A. Fraser.
\newblock 1993.
\newblock Black box and glass box evaluation of the {SUNDIAL} system.
\newblock In {\em Proceedings of the Third European Conference on Speech
  Communication and Technology}, pages 1423--1426.

\bibitem[\protect\citename{Smith and Gordon}1997]{SmithGordon97}
Smith, Ronnie~W. and Steven~A. Gordon.
\newblock 1997.
\newblock Effects of variable initiative on linguistic behavior in
  human-computer spoken natural language dialog.
\newblock {\em Computational Linguistics}, 23(1).

\bibitem[\protect\citename{Sparck-Jones and Galliers}1996]{sjg96}
Sparck-Jones, Karen and Julia~R. Galliers.
\newblock 1996.
\newblock {\em Evaluating Natural Language Processing Systems}.
\newblock Springer.

\bibitem[\protect\citename{Walker}1996]{Walker96a}
Walker, Marilyn~A.
\newblock 1996.
\newblock {The Effect of Resource Limits and Task Complexity on Collaborative
  Planning in Dialogue}.
\newblock {\em {Artificial Intelligence Journal}}, 85(1--2):181--243.

\bibitem[\protect\citename{Webber and Joshi}1982]{WJ82}
Webber, Bonnie and Aravind Joshi.
\newblock 1982.
\newblock Taking the initiative in natural language database interaction:
  Justifying why.
\newblock In {\em Coling 82}, pages 413--419.

\end{thebibliography}

\end{document}